\documentclass{article}

\usepackage[preprint]{neurips_2025}

\usepackage{microtype}
\usepackage{hyperref}
\usepackage{url}
\usepackage{booktabs}
\usepackage{graphicx}
\usepackage{amsfonts}       %
\usepackage{nicefrac}  
\usepackage{multirow}
\usepackage{multicol}
\usepackage{color,xcolor}
\usepackage[caption=false]{subfig}
\usepackage{booktabs}
\usepackage{threeparttable}
\usepackage{courier}
\usepackage{adjustbox}
\usepackage{supertabular}
\usepackage{stmaryrd}
\usepackage{makecell}
\usepackage{marvosym}
\usepackage{colortbl}
\usepackage{pifont}
\usepackage{amsmath} 
\colorlet{darkgreen}{green!65!black}
\colorlet{darkred}{red!80!black}
\definecolor{lightblue}{HTML}{0071bc}
\definecolor{lightgreen}{HTML}{39b54a}
\usepackage{caption}
\usepackage{times}
\usepackage{epsfig}
\usepackage{graphicx}
\usepackage{verbatim}
\usepackage{bbm}
\usepackage{wrapfig}
\usepackage{subcaption}

\usepackage{subcaption}

\usepackage{enumitem,kantlipsum}

\newcolumntype{g}{>{\columncolor{tbgray}}c}

\definecolor{red}{rgb}{0.8, 0.0, 0.0}
\definecolor{green}{rgb}{0.0, 0.5, 0.0}
\definecolor{tbgray}{gray}{.92}

\usepackage{lineno}

\definecolor{darkblue}{rgb}{0, 0, 0.5}
\hypersetup{colorlinks=true, citecolor=darkblue, linkcolor=darkblue, urlcolor=darkblue}

\title{Breaking Down Video LLM Benchmarks: Knowledge, Spatial Perception, or True Temporal Understanding?}

\author{Bo Feng$^\circ$, Zhengfeng Lai$^\star$$^\circ$, Shiyu Li$^\circ$, Zizhen Wang$^\circ$ \\ \textbf{Simon Wang}, \textbf{Ping Huang$^\dagger$}, \textbf{Meng Cao$^\dagger$}  \\
Apple \\
\texttt{\{bfeng2, jeff\_lai, shiyu\_li, wang\_zizhen\}@apple.com} \\
\texttt{\{simon\_wang2, huang\_ping, mengcao\}@apple.com} \\
$^\circ$Equal contributions; $^\star$Corresponding author;$^\dagger$Senior authors}

\begin{document}

\maketitle

\begin{abstract}
Existing video understanding benchmarks often conflate knowledge-based and purely image-based questions, rather than clearly isolating a model's temporal reasoning ability, which is the key aspect that distinguishes video understanding from other modalities. We identify two major limitations that obscure whether higher scores truly indicate stronger understanding of the dynamic content in videos: (1) strong language priors, where models can answer questions without watching the video; and (2) shuffling invariance, where models maintain similar performance on certain questions even when video frames are temporally shuffled. To alleviate these issues, we propose VBenchComp, an automated pipeline that categorizes questions into different domains: LLM-Answerable, Semantic, and Temporal. Specifically, LLM-Answerable questions can be answered without viewing the video; Semantic questions remain answerable even when the video frames are shuffled; and Temporal questions require understanding the correct temporal order of frames. The rest of the questions are labeled as Others. This can enable fine-grained evaluation of different capabilities of a video LLM. 
Our analysis reveals nuanced model weaknesses that are hidden by traditional overall scores, and we offer insights and recommendations for designing future benchmarks that more accurately assess video LLMs.

\end{abstract}

\section{Introduction}

% \begin{table}[h]
% \centering
% \caption{GPU hours needed for evaluating video LLMs across different benchmarks.}
% \vspace{0.2cm}

% \begin{wraptable}{r}{0.6\textwidth}
% \vspace{-10pt}
% \centering
% \small
% \resizebox{0.60\textwidth}{!}{
% \small
% \begin{tabular}{lc|ccc}
% \caption{GPU hours needed for evaluating video LLMs across different benchmarks.}
% \vspace{0.2cm}
% \toprule
% \multirow{2}{*}{Benchmark} & \multirow{2}{*}{\begin{tabular}[c]{@{}l@{}}Number of \\ Questions\end{tabular}} & \multicolumn{3}{c}{Model Size} \\
%                            &                                                                                 & 2B       & 7B       & 72B      \\
% \midrule
% LongVideoBench~\cite{wu2024longvideobench} & 1337    & 14.3     & 15.0     & 26.4    \\
% Egoschema~\cite{mangalam2024egoschema} & 500 & 2.7 & 2.2 & 6.7 \\
% NexTQA~\cite{xiao2021next} & 4996 & 16.0 & 20.0 & 58.0 \\
% VideoMME~\cite{fu2024video} & 2700 & 18.0 & 20.7 & 50.7 \\
% MLVU~\cite{mlvu} & 2174 & 14.1 & 15.9 & 31.7 \\
% LVBench~\cite{lvbench} & 1549 & 7.5 & 8.7 &22.2 \\
% PerceptionTest~\cite{perceptiontest} & 19140 & 118.1 & 131.2 & 296.3 \\
% % \midrule

% \bottomrule
% \label{tab: gpu_hour}
% \end{tabular}

% % \vspace{-0.5cm}
% }
% \end{wraptable}
% % \end{table}

\begin{wraptable}{r}{0.6\textwidth}
\vspace{-10pt}
\centering
\small
\caption{A100 GPU hours needed for evaluating video LLMs (Qwen2-VL) across different benchmarks.}
\resizebox{0.60\textwidth}{!}{
\begin{tabular}{lc|ccc}
\toprule
\multirow{2}{*}{Benchmark} & \multirow{2}{*}{\begin{tabular}[c]{@{}l@{}}Number of \\ Questions\end{tabular}} & \multicolumn{3}{c}{Model Size} \\
                           &                                                                                 & 2B       & 7B       & 72B      \\
\midrule
LongVideoBench~\cite{wu2024longvideobench} & 1337    & 14.3     & 15.0     & 26.4    \\
Egoschema~\cite{mangalam2024egoschema} & 500 & 2.7 & 2.2 & 6.7 \\
NexTQA~\cite{xiao2021next} & 4996 & 16.0 & 20.0 & 58.0 \\
VideoMME~\cite{fu2024video} & 2700 & 18.0 & 20.7 & 50.7 \\
MLVU~\cite{mlvu} & 2174 & 14.1 & 15.9 & 31.7 \\
LVBench~\cite{lvbench} & 1549 & 7.5 & 8.7 &22.2 \\
PerceptionTest~\cite{perceptiontest} & 19140 & 118.1 & 131.2 & 296.3 \\
\midrule
Total & 32396 & 190.6 & 213.7 & 491.9 \\

\bottomrule
\end{tabular}
}
\label{tab: gpu_hour}
\vspace{-8pt}
\end{wraptable}

The rapid progress of video Large Language Models (video LLMs) has led to the emergence of a wide range of video understanding benchmarks, such as VideoMME~\cite{fu2024video}, MLVU~\cite{mlvu}, LongVideoBench~\cite{wu2024longvideobench}, EgoSchema~\cite{mangalam2024egoschema}, and others. While this surge of benchmarks offers broader coverage for evaluating different capabilities, it also introduces considerable computational cost and redundancy. As shown in Table~\ref{tab: gpu_hour}, evaluating a 2B-parameter model (e.g., Qwen2-VL) across existing video QA benchmarks requires 190.6 A100 GPU hours. This computational cost escalates dramatically to 491.9 hours for a 72B model, raising serious concerns about the computational burden of benchmarking video LLMs given the growing number of video understanding datasets.

\begin{figure*}[!t]
    \centering
    \includegraphics[width=0.9\linewidth]{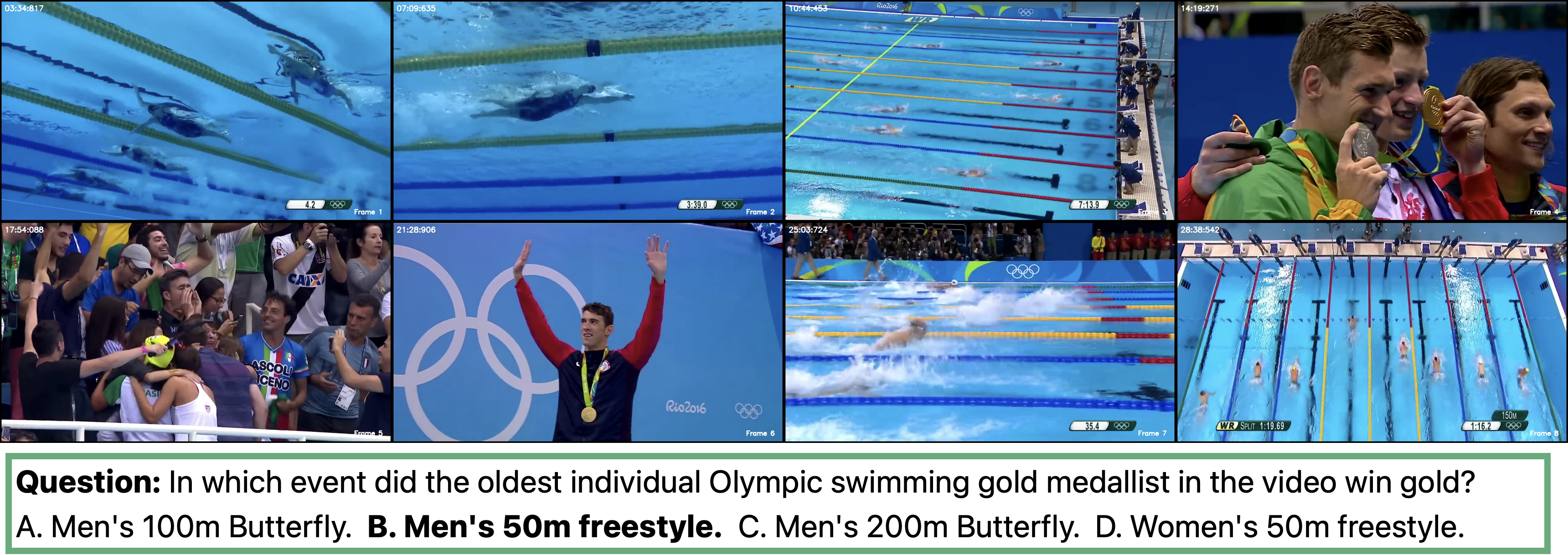} \\
    \vspace{0.2cm}
    \includegraphics[width=0.9\textwidth]{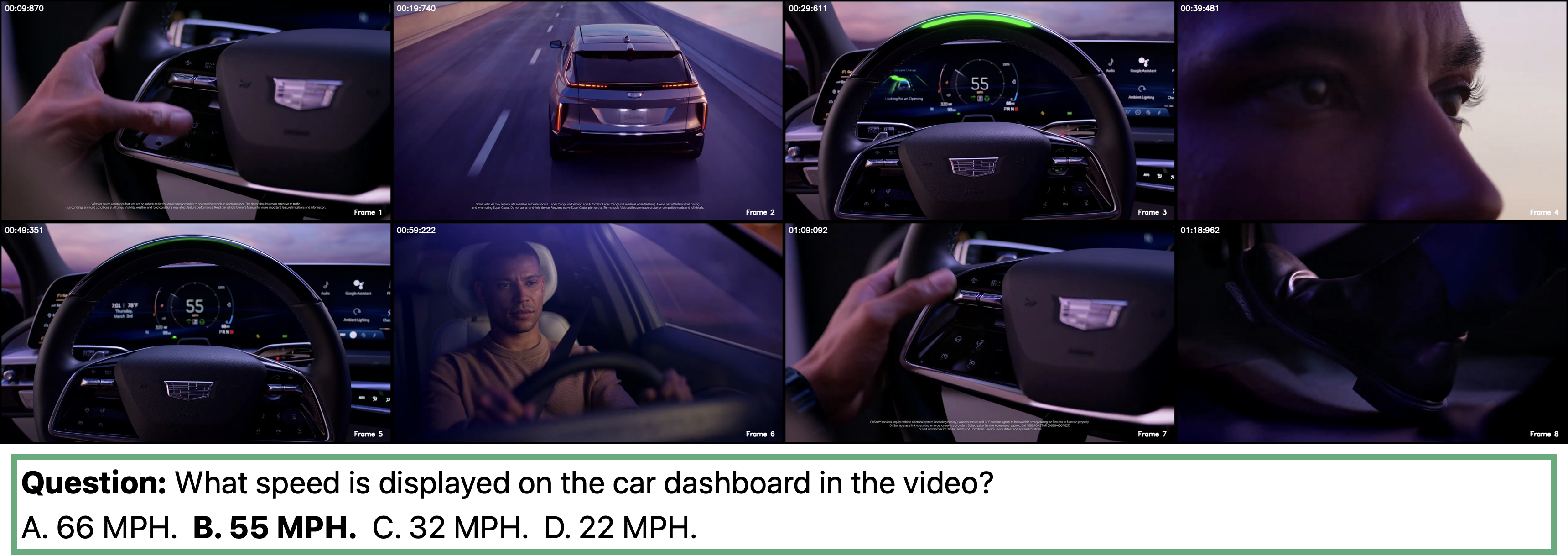} \\
    \vspace{0.2cm}
    \includegraphics[width=0.9\textwidth]{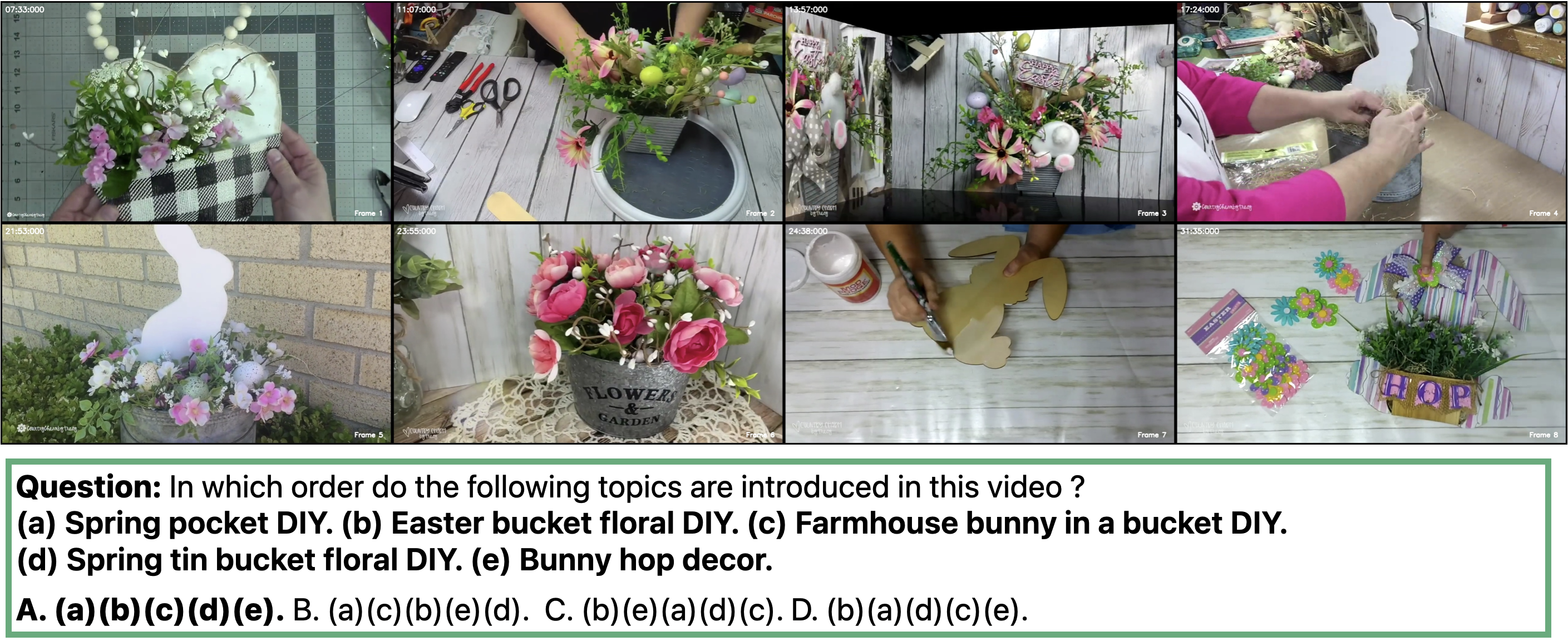}
    \caption{\small{Examples of LLM-Answerable, Semantic and Temporal questions in VideoMME~\citep{fu2024video}: (Top) The model uses LLM's prior knowledge to answer correctly without the need of video; (Middle) The model relies on semantic understanding to answer without requiring temporal comprehension; (Bottom) The model relies on comprehensive temporal understanding to answer. }}
    \vspace{-0.5cm}
    \label{fig:text_only}
\end{figure*}

Beyond the computational cost, current video understanding benchmarks often conflate different skills and fail to truly evaluate the video understanding capability. 
We identify two key limitations that undermine meaningful evaluation. First, some questions can be answered correctly without access to the video, since models rely on their pretrained language priors rather than visual evidence, as shown in Figure~\ref{fig:text_only}. These questions primarily test the underlying LLM’s factual knowledge and reasoning skills, rather than evaluating the model’s ability to process and understand visual content. As a result, high performance on these questions can misleadingly inflate benchmark scores, giving the false impression of strong video understanding when, in fact, the model may not be attending to the visual input at all. Second, some questions primarily assess static semantic understanding and do not require comprehension of the video's temporal structure. For example, models often achieve similar performance even when the video frames are randomly shuffled, indicating that their predictions rely heavily on spatial or frame-level cues rather than temporal reasoning. This shuffling invariance exposes a critical flaw: current benchmarks may significantly overestimate a model’s true temporal understanding, conflating static visual recognition with dynamic sequence reasoning.

While many existing benchmarks claim to be comprehensive, there is currently no standardized protocol for assessing their effectiveness. 
Each dataset emphasizes different aspects of video comprehension, yet lacks a clear metric for how well it captures temporal reasoning, which is the core capability that distinguishes video from static images. We introduce VBenchComp, an automated evaluation pipeline that categorizes questions into four distinct domains: LLM-Answerable, Semantic, Temporal, and Other. This structured categorization disentangles the contributions of language priors, static visual understanding, and genuine temporal reasoning, enabling a more diagnostic and interpretable evaluation of video models. 
Based on this,  we curate a core benchmark subset that emphasizes both semantic and temporal understanding, and introduce a dedicated metric, the VBenchComp Score, which provides a more focused and light-weighted evaluation protocol to better guide model development and comparison. Importantly, we find that results obtained from this core set are consistent with those from the full benchmark suite, while reducing computational cost significantly.
 
\section{Related Works}

\textbf{Video Large Language Models (Video LLMs).} 
Large Language Models (LLMs) have revolutionized natural language understanding, demonstrating exceptional ability to follow human instructions and serving as versatile agents for general-purpose AI assistants~\citep{chowdhery2023palm,touvron2023llama2,achiam2023gpt}.
Building on these advancements, Multimodal Large Language Models (MLLMs)~\citep{liu2023improvedllava, liu2024llavanext} have made significant strides in vision-language learning by incorporating visual encoders with LLMs and fine-tuning on vision-language instruction data. 
In addition, video Large Language Models (video LLMs) incorporate visual encoders to extract video features, temporal modeling mechanisms to capture motion dynamics, and large language models to generate responses~\citep{chen2024fewer}. For instance, Video-ChatGPT~\citep{Maaz2023VideoChatGPT} employs CLIP~\citep{radford2021learning} to obtain per-frame representations, which are subsequently processed through spatial and temporal pooling before being fed into an LLM. LLaVA-NeXT-Video~\citep{zhang2024llavanextvideo} builds on LLaVA-NeXT~\citep{liu2024llavanext}, adapting it for video-based tasks, while its DPO-enhanced variant~\citep{zhang2024llavanextvideo} further refines output quality by aligning responses with AI-generated feedback.
To improve temporal consistency, VideoLLaMB~\citep{wang2024videollamb} incorporates memory tokens within its bridge layers, allowing the model to capture both sequential dependencies and historical visual context. InternVideo2.5~\citep{internvideo2.5} enhances multimodal models by leveraging annotations from dense vision tasks, optimizing preferences directly, and refining spatiotemporal representations through hierarchical token compression. This enables better handling of detailed video content and extended temporal reasoning.
Apollo~\citep{zohar2024apollo} introduces a structured methodology for training video LLMs, while Oryx~\citep{liu2024oryx} proposes OryxViT, a vision transformer pre-trained to encode images at arbitrary resolutions into representations compatible with LLMs. Oryx also integrates a dynamic compression module that adjusts visual token density, allowing compression levels from 1x to 16x based on task requirements.
VideoLLaMA-3~\citep{zhang2025videollama3} highlights the importance of high-quality image-text data for both image and video comprehension. 

\textbf{Video Understanding Benchmarks.} Traditional evaluations of video LMMs rely on classic video QA benchmarks such as MSRVTT-QA~\citep{xu2017video} and ActivityNet-QA~\citep{yu2019activityqa}, which assess global video understanding through summary-based questions. However, prior work has shown that these datasets can often be answered using only a handful of key frames, limiting their effectiveness in evaluating true temporal reasoning~\citep{wu2024longvideobench}. More recent benchmarks, such as NeXT-QA~\citep{xiao2021next} and MVBench~\citep{li2024mvbench}, focus on short clips (averaging 44 and 16 seconds, respectively), while Video-MME~\citep{fu2024video} spans a diverse set of video domains and durations. LongVideoBench~\citep{wu2024longvideobench} explicitly targets referring reasoning over long videos. While each benchmark aims to provide a comprehensive evaluation of video understanding, there is still no standardized protocol to assess their effectiveness—particularly in measuring temporal reasoning. A rigorous framework is needed to systematically evaluate these benchmarks, ensuring that they go beyond static frame-based understanding and capture the core challenges of video comprehension.
\section{A Closer Look at Video Understanding Benchmarks}

Evaluating the video understanding capabilities of recent video LLMs is a complex and multifaceted task. Although many benchmarks exist,  their ability to capture the depth of reasoning needed for real-world video comprehension remains uncertain. Some focus on higher-level skills like temporal reasoning, causal inference, and fine-grained event recognition, while others may primarily focus on semantic understanding. To explore this, we take a closer look at several widely used video understanding benchmarks. We select VCGBench~\citep{Maaz2023VideoChatGPT} and ActivityNet-QA~\citep{yu2019activityqa} as representative open-ended QA benchmarks. We use  NeXT-QA~\citep{xiao2021next}, VideoMME~\citep{fu2024video}, EgoSchema~\citep{mangalam2024egoschema}, MLVU~\citep{mlvu}, and LongVideoBench~\citep{wu2024longvideobench} to serve as examples of multiple-choice QA benchmarks.

% \begin{figure}[t]
%     \centering
%     \includegraphics[width=0.99\linewidth]{Figure/text_exp1.png}
%     % \vspace{-0.3cm}
%     \caption{\small{An example of LLM-Answerable Question in VideoMME~\citep{fu2024video}: the model can use LLM's prior knowledge to answer the question correctly without watching the video.}}
%     \label{fig:text_only}
%     \vspace{-0.2cm}
% \end{figure} 

% \textbf{video LLMs can answer questions without videos. }  

\begin{wrapfigure}{r}{0.48\textwidth}
\vspace{-0.3cm}
    \centering
    \includegraphics[width=0.48\textwidth]{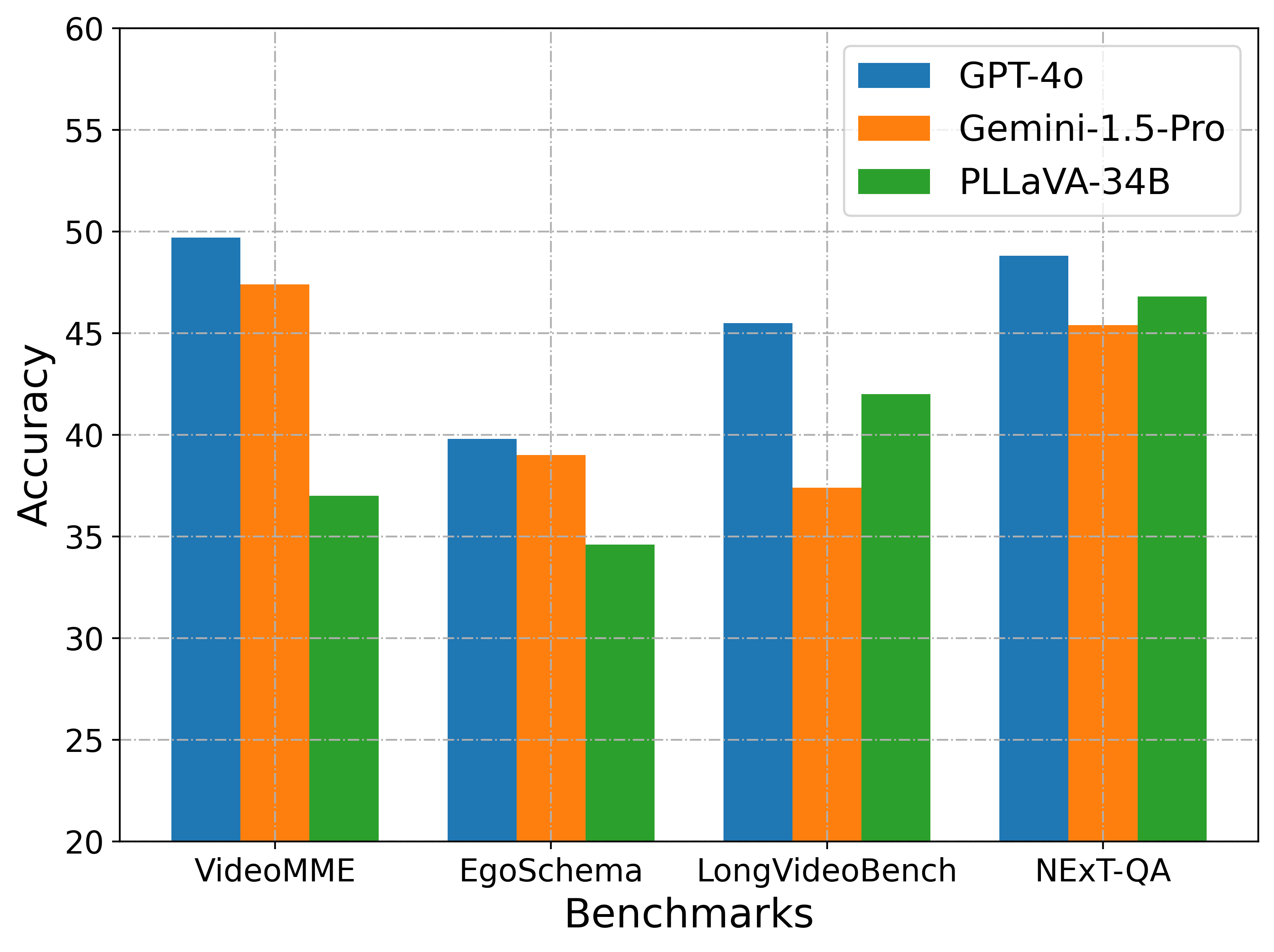}
    \vspace{-0.2cm}
    \caption{\small{Performance of different MLLMs without videos as the input on four benchmarks.}}
    \vspace{-0.5cm}
    \label{fig:llm_answerable}   
\end{wrapfigure}
\subsection{Answering Questions without Videos}

Inspired by \cite{zhu2024lime}, we begin by evaluating whether questions can be answered without access to the corresponding videos. To do this, we input the questions into various MLLMs without providing the videos. As shown in Figure~\ref{fig:llm_answerable}, surprisingly, GPT-4o achieves up to 50\% accuracy on both VideoMME and NExT-QA, despite not processing any video data. Similarly, open-sourced models like PLLaVA-34B also achieve 37.0\% accuracy on VideoMME without video input. For long video understanding (LongVideoBench~\cite{wu2024longvideobench}), these models even surpass 35\% accuracy without feeding in the actual long videos. These results cast serious doubt on the reliability of these benchmarks in accurately assessing a model's video understanding capabilities. The fact that models can achieve notable performance without processing the video data suggests that these benchmarks may be evaluating factors unrelated to true video comprehension—such as reliance on text-based cues or prior knowledge. This raises the question of whether these benchmarks are an effective measure of multimodal models' abilities to understand and process video content. Figure~\ref{fig:text_only} shows an example that the question can be directly answered by an LLM without watching the video.

\begin{figure*}[!t]
\begin{center}
	\begin{tabular}{c}
	\includegraphics[width=0.96\textwidth]{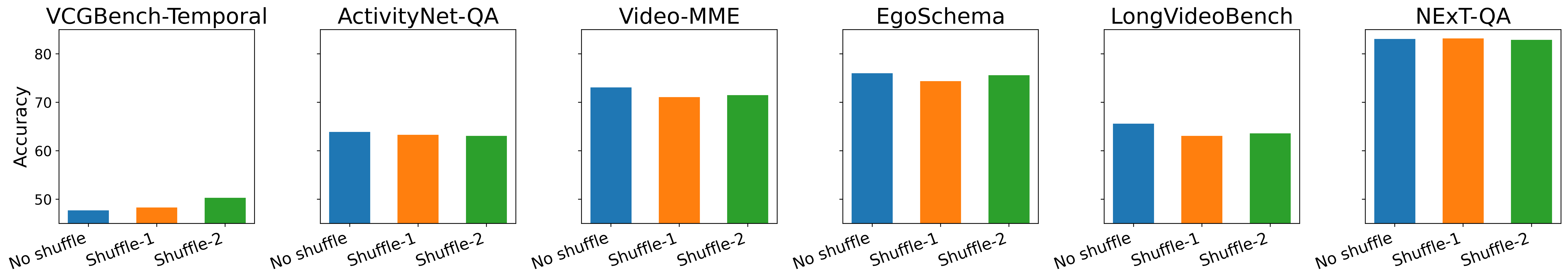}  \\
        \footnotesize{(a) Close-sourced GPT-4o}  \\
 % (a) CLIP & (b) VeCLIP \\
 
	% \includegraphics[width=0.96\textwidth]{Figure/Gemini-1.5-Pro.png}    \\
	  % \footnotesize{(b) Close-sourced Gemini-1.5-Pro}  \\  % (c) CLIP  & (d) VeCLIP \\
        \includegraphics[width=0.96\textwidth]{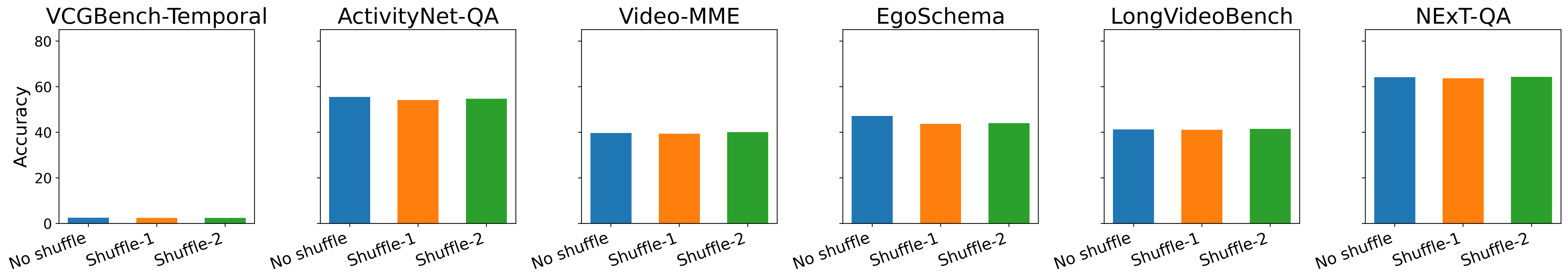}    \\
	  \footnotesize{(b) Training-free Model: SlowFast-LLaVA-7B~\citep{xu2024slowfast}}  \\
        \includegraphics[width=0.96\textwidth]{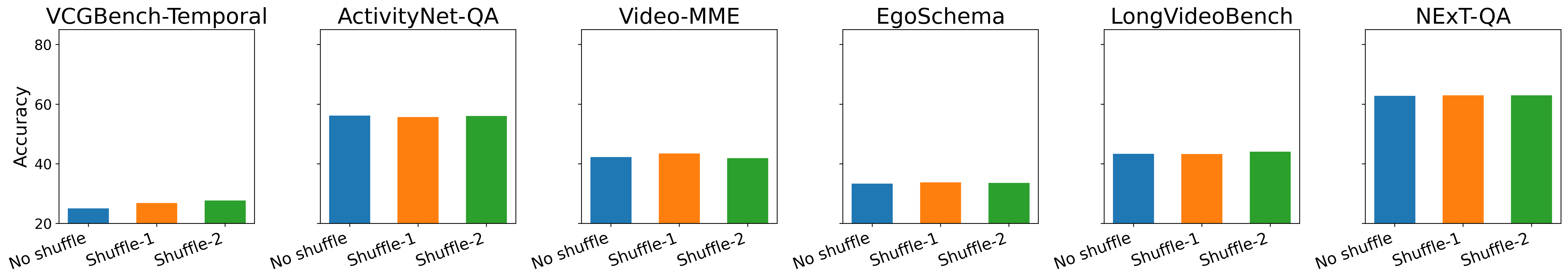}    \\
	  \footnotesize{(c) LORA Fine-tuned Model: PLLaVA-7B~\citep{xu2024pllava}}  \\ 
   %      \includegraphics[width=0.96\textwidth]{Figure/PLLaVA-34B.png}    \\
	  % \footnotesize{(c) LORA Fine-tuned Model: PLLaVA-34B}  \\ 
        \includegraphics[width=0.96\textwidth]{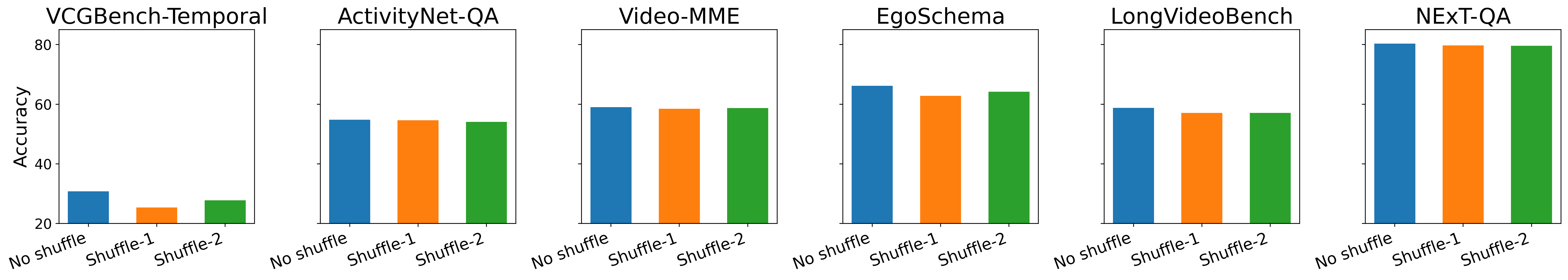}    \\
	  \footnotesize{(d) Video SFT Model: LLaVA-OV-7B-Qwen2~\citep{li2024llava}}  \\
   %      \includegraphics[width=0.96\textwidth]{Figure/LLaVA-OV-72B-Qwen2.png}    \\
	  % \footnotesize{(e) Video SFT Model: LLaVA-OV-72B-Qwen2~\citep{li2024llava}}  \\
   %      \includegraphics[width=0.96\textwidth]{Figure/Qwen2-VL-7B.png}    \\
	  % \footnotesize{(g) Video SFT Model: Qwen2-VL-7B}  \\
	\end{tabular}
\end{center}
% \vspace{-0.5cm}
\caption{\small{After shuffling the extracted frames, the scores of each model remain unshaken across all benchmarks. *Frame settings: (a), (d) uses 128 frames for VideoMME-long, others use 64 frames; (b) uses \(10_{\text{slow}} + 50_{\text{fast}}\) frames for all benchmarks; (c) uses 16 frames for all benchmarks. }}
\vspace{-0.5cm}
\label{fig:shuffled_frames}
\end{figure*}

\subsection{Shuffled Frames but Unshaken Scores}
Besides the aforementioned issues on the text, we also identify a critical concern regarding the semantic and temporal understanding. In recent video LLMs, multiple frames are typically sampled and fed into the model. However, many questions in these benchmarks may not adequately assess a model's ability to understand the temporal dynamics of videos. A natural approach to evaluating temporal understanding would be to test whether shuffling the frames affects the model's final answer. If a model can still produce accurate responses despite shuffled frames, it suggests that the question may not require a deep understanding of temporal relationships, but rather relies on static or semantic content from the frames themselves. We conduct experiments with a variety of representative video LLMs across different model types: 1) closed-source models such as GPT-4o and Gemini-1.5-Pro~\citep{team2024gemini}, which set the state-of-the-art standard for video understanding; 2) training-free models such as SlowFast-LLaVA~\citep{xu2024slowfast}, which leverage pre-trained visual and language models without additional fine-tuning; 3) LoRA fine-tuned models, e.g., PLLaVA~\citep{xu2024pllava}, which demonstrate the effectiveness of parameter-efficient adaptation; and 4) video SFT models, such as LLaVA-OneVision (LLaVA-OV)~\citep{li2024llava}, which benefit from supervised fine-tuning with video datasets. This selection enables us to systematically assess the effect of shuffled frames on a wide range of models. 

As shown in Figure~\ref{fig:shuffled_frames}, we apply frame shuffling twice and observe that the scores of both GPT-4o and Gemini-1.5-Pro remain remarkably stable, indicating that these models are largely unaffected by temporal disruptions. A similar pattern is observed in open-source models such as SlowFast-LLaVA, PLLaVA, and LLaVA-OV, despite differences in their training paradigms and architectures. Interestingly, this insensitivity to temporal order persists across models of varying sizes. For instance, the large model (LLaVA-OV-72B-Qwen2) and the smaller model (LLaVA-OV-7B-Qwen2) exhibit consistent behavior across all six benchmarks. Surprisingly, in some cases, shuffling the frames even leads to an improvement in performance. For example, Gemini-1.5-Pro achieves a higher score on EgoSchema after frame shuffling, and GPT-4o also performs better on NExT-QA under the same condition. This counter-intuitive result raises critical concerns about the validity of these benchmarks in evaluating true video understanding. If models can achieve higher scores after the temporal sequence is disrupted, it suggests that these benchmarks may not be adequately assessing the models' ability to comprehend and reason over temporal information, which is a core component of video understanding. 

\subsection{Potentially Misleading Scores in Current Video Benchmarks} 
% Our analysis reveals that a single final score reported by current video benchmarks may not accurately reflect a model's true capability in video understanding. A closer examination suggests that many tasks within these benchmarks are overly reliant on prior language knowledge and semantic understanding, allowing models to excel without genuinely engaging with the temporal aspects of the video. Specifically, models often exploit spatial correlations and semantic associations across individual frames, bypassing the need to process or reason about temporal dependencies.
Based on the above analysis, we find that a single final score reported by current video benchmarks may not accurately reflect a model's true capability in video understanding: many of the tasks may be overly relying on prior language knowledge or semantic understanding rather than requiring genuine video understanding across dynamic frames. As a result, models might excel by leveraging spatial correlations and semantic associations within individual frames, bypassing the need to process temporal dependencies. This raises the possibility that benchmark scores could be misleading, potentially leaving the impression that models possess a more profound understanding of video content than they actually do. Moreover, these results suggest that current video benchmarks inadvertently prioritize evaluating the LLM's language proficiency and semantic understanding over its temporal comprehension of video content. This overemphasis can lead to biased evaluations, where models with strong language priors or frame-level understanding receive inflated scores, despite having limited capability to capture complex temporal dynamics. Such biases introduce the risk of drawing erroneous conclusions about a model's progress in video understanding, potentially giving a false sense of achievement in the field thus making community risk overestimating the robustness and real-world applicability of these models.
% Without benchmarks that comprehensively test for temporal coherence, causal reasoning, and action understanding over time, the community risks overestimating the robustness and real-world applicability of these models.
Therefore, we advocate for the development of more comprehensive evaluation protocols that disentangle language knowledge, semantic, and temporal understanding, ensuring a more accurate and holistic assessment of video models.

\section{A Standardized Protocol for Breaking Down Video LLM Benchmarks}

\begin{figure}[!t]
    \centering
    \includegraphics[width=0.99\linewidth]{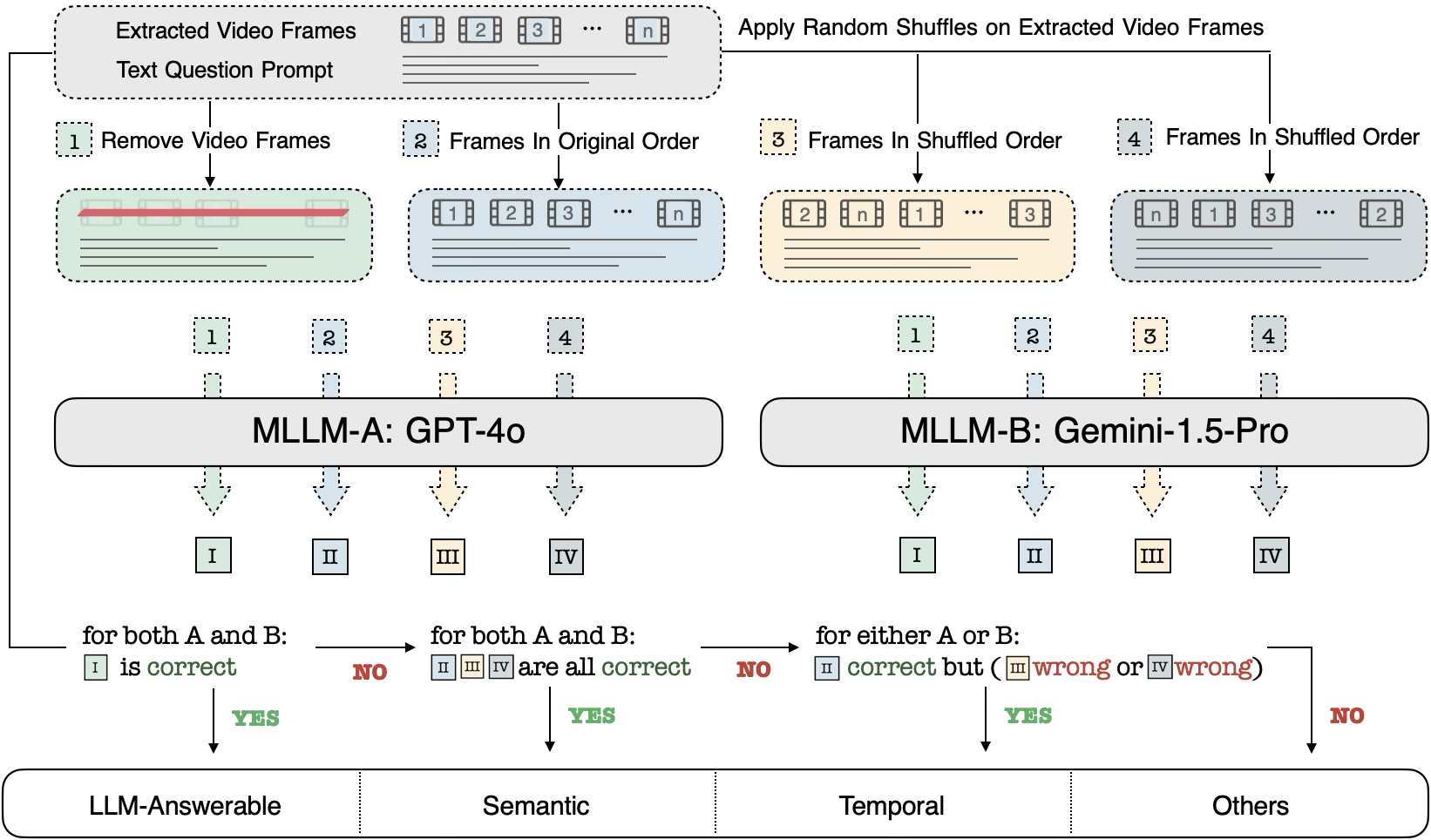}
    % \vspace{-0.3cm}
    \caption{\small{An overview of our standardized protocol: benchmark questions are categorized into four groups. Questions answerable by both GPT-4o and Gemini without video are classified as LLM-Answerable. For the remaining questions, we apply random shuffles to the extracted frames twice: if both models answer correctly before and after shuffling, the question is classified as Semantic. If one model answers correctly before but fails after shuffling, the question is classified as Temporal. All other questions are categorized as Others.}}
    \label{fig:pipeline}
    \vspace{-0.2cm}
\end{figure} 

In this section, we propose a standardized protocol (as shown in Figure~\ref{fig:pipeline}) for decomposing video LLM benchmarks into four distinct domains: (1) LLM-answerable questions to focus on the prior language capabilities of the LLM backbone, (2) semantic understanding questions to evaluate the model's ability to understand semantic content, (3) temporal understanding questions to measure the model’s capacity to capture temporal dependencies and dynamic changes, and (4) Other questions that may either require overly advanced comprehensive reasoning or are poorly constructed and thus lack sufficient distinctiveness.  Our goal is to disentangle these question types to provide a more precise and comprehensive evaluation of video LLMs. It will also assist future benchmarks in refining their question design strategies and focusing more on authentic video-understanding questions.

\subsection{LLM-Answerable Questions}
Answer leakage is a critical issue in image-QA benchmarks, where MLLMs can often generate correct answers without relying on the image itself. Instead of genuinely integrating visual and textual information, these models leverage their pre-trained knowledge from LLM to infer answers based solely on the text~\citep{zhu2024lime}. This undermines the intended goal of evaluating a model’s multimodal understanding capabilities. Multimodal answer leakage can be summarized into two categories: 1) text-answerable questions, where the question itself provides sufficient information for the model to answer, rendering the associated visual input unnecessary; 2) memorized questions, where the MLLM has previously encountered the same question during training and recalls the corresponding answer from memory rather than reasoning from the given image. As a result, certain questions can be answered solely by a text-based LLM without requiring visual input. To assess this, we perform a text-only evaluation using both GPT-4o and Gemini-1.5-Pro. As shown in Figure~\ref{fig:pipeline}, if both models correctly answer a given question without the video, we classify the corresponding QA pair as an LLM-Answerable question. We then analyze the entire benchmark and compute the proportion of such questions relative to the total, denoted as $\alpha$.

\subsection{Semantic Questions: Shuffling Frames but Consistent Answer}
After filtering for LLM-answerable questions, we further identify a subset of questions that focus specifically on semantic understanding. To achieve this, we introduce a diagnostic procedure: for each video-question pair, we first generate answers using Gemini-1.5-Pro and GPT-4o. We then shuffle the extracted frames and query the models again - repeating this process twice. If both models consistently provide correct answers despite the disrupted temporal order (before and after shuffling the extracted frames), we classify the question as semantic, indicating that static visual information from a single or a certain group of frames alone are sufficient for answering.
By applying this procedure across the benchmark, we compute the proportion of such questions, denoted as $\beta$, to quantify the prevalence of questions relying solely on semantic understanding. A high $\beta$ suggests that the benchmark may be biased toward spatial or appearance-based cues, potentially inflating a model’s perceived temporal reasoning capability. This highlights the need to construct more temporal-related questions that explicitly require sequential understanding to ensure a more rigorous and targeted evaluation of video LLMs.

\subsection{Temporal Questions}

After classifying questions into LLM-Answerable and Semantic categories, the remaining questions are further divided into Temporal and Others. To identify Temporal questions, we apply the following criterion: if GPT-4o or Gemini-1.5-Pro answers the question correctly when provided with frames in their original order but fails to do so after the frames are shuffled, we classify the question as Temporal, indicating that the right sequential information is crucial for the answering process. Unlike semantic or frame-independent tasks, these questions assess whether the model can correctly infer event progression and temporal consistency over time. By introducing a controlled perturbation—shuffling the frame order, we isolate the questions for temporal understanding capacity, distinguishing them from purely visual or semantic understanding.

\subsection{Others}

Lastly, the remaining questions will be labeled as Others. This category includes questions that are either too difficult to answer for all SOTA models or are so comprehensive that they may require additional modalities, such as audio, to resolve. Questions may depend on recognizing spoken dialogue, distinguishing between environmental sounds, or interpreting non-visual context cues like tone or timing. For example, in VideoMME~\citep{fu2024video}, answering certain questions may depend on recognizing spoken dialogue, distinguishing between environmental sounds, or interpreting non-visual context cues like tone or timing.

\subsection{VBenchComp: Quantifying Video Benchmark Composition}
\label{sec:vbenchcomp}
To systematically analyze and quantify the composition of video LLM benchmarks, we introduce \textbf{VBenchComp}, a diagnostic tool that applies our standardized protocol (Figure~\ref{fig:pipeline}) to decompose the benchmark into its four key domains. VBenchComp computes the ratios of LLM-Answerable, Semantic, Temporal, and Others questions, denoted as $\alpha$, $\beta$, $\gamma$, and $\delta$ respectively.

\paragraph{Benchmark profiling and skill gap identification.} VBenchComp not only quantifies benchmark composition but also identifies potential gaps in coverage. For instance, an overrepresentation of LLM-Answerable questions ($\alpha$) suggests that the benchmark may underestimate the need for genuine multimodal understanding. Conversely, an excess of Semantic questions ($\beta$) could create an illusion of strong temporal understanding, when in reality, the model might rely primarily on static frame information. A low proportion of Temporal questions ($\gamma$) may indicate inadequate assessment of dynamic event comprehension. 

\section{Experimental Results}
\subsection{An Overview of VBenchComp}
We apply the standardized categorization protocol described in Section~\ref{sec:vbenchcomp} to seven widely-used video question answering benchmarks, quantifying their distributions across four diagnostic categories: LLM-Answerable, Semantic, Temporal, and Others. Table~\ref{tab: composition} summarizes the raw counts and their corresponding percentages relative to the total number of questions in each benchmark. Across all benchmarks, we observe a considerable variation in the proportion of question types, which reflects their differing emphases on language, semantic, and temporal capabilities. For instance, \textit{NextQA}~\cite{xiao2021next}, \textit{LongVideoBench}~\cite{wu2024longvideobench}, \textit{MLVU}~\cite{mlvu}, \textit{Egoschema}~\cite{mangalam2024egoschema}, and \textit{VideoMME}~\cite{fu2024video}  contain a significant portion of LLM-Answerable questions, which indicates potential answer leakage and reliance on language priors. In contrast, benchmarks like \textit{LVBench}~\cite{lvbench} contains relatively fewer LLM-Answerable questions. On the other hand, with the exception of \textit{LongVideoBench} and \textit{LVBench}, all other benchmarks have more than 30\% of Semantic questions, where frame shuffling has minimal impact on the model’s ability to produce correct answers. 

\begin{table}[!t]
\centering
\caption{Compositions of question types across different video understanding benchmarks. Each cell (except Total) shows the count and its percentage of the total.}
\vspace{0.2cm}
\small
\begin{tabular}{lccccc}
\toprule
\textbf{Dataset} & \textbf{Total} & \textbf{Text} & \textbf{Semantic} & \textbf{Temporal} & \textbf{Others} \\
\midrule
LongVideoBench~\cite{wu2024longvideobench} & 1337 & 363 / 27.15\% & 308 / 23.03\% & 235 / 17.58\% & 431 / 32.24\% \\
Egoschema~\cite{mangalam2024egoschema}      & 500  & 133 / 26.60\% & 182 / 36.40\% & 45 / 9.00\%   & 140 / 28.00\% \\
NextQA~\cite{xiao2021next}         & 4996 & 1738 / 34.79\% & 1880 / 37.63\% & 437 / 8.75\%  & 941 / 18.83\% \\
VideoMME~\cite{fu2024video}       & 2700 & 841 / 31.15\% & 810 / 30.00\% & 371 / 13.74\% & 678 / 25.11\% \\
MLVU~\cite{mlvu}           & 2174 & 621 / 28.57\% & 643 / 29.57\% & 383 / 17.62\% & 527 / 24.23\% \\
LVBench~\cite{lvbench}        & 1549 & 140 / 9.04\%  & 321 / 20.72\% & 355 / 22.92\% & 733 / 47.32\% \\
PerceptionTest~\cite{perceptiontest} & 19140 & 3642 / 19.03\% & 6283 / 32.82\% & 3117 / 16.29\% & 6098 / 31.86\% \\
\bottomrule
\end{tabular}
\label{tab: composition}
\vspace{-0.5cm}

\end{table}

\begin{table}[h]
\caption{\centering
Benchmarking public models under VBenchComp categorization. (All settings use 64 frames, except for VideoMME-long, which uses 128.)  }
\hspace{0.3cm}
    \centering
    \begin{tabular}{c@{\hskip 0.1in}c}

           \small (a) Egoschema~\cite{mangalam2024egoschema} & \small (b) NextQA~\citep{xiao2021next} \\
        % First Table
        \resizebox{0.47\textwidth}{!}{
        \begin{tabular}{cc|c|cccc}
            \toprule
Size                 & Model       & Overall & LLM  & Semantic & Temporal & Others \\
\midrule
\multirow{3}{*}{7B}  & Qwen2-VL~\cite{wang2024qwen2}    & 65.8    & 85.0 & 83.5     & 37.8     & 33.6   \\
                     & LLaVA-OV~\cite{li2024llava}    & 66.2    & 75.2 & 83.5     & 57.8     & 37.9   \\
                     & LLaVA-Video~\cite{lin2023video} & 61.8    & 72.2 & 82.4     & 46.7     & 30.0   \\
                     \midrule
\multirow{3}{*}{72B} & Qwen2-VL~\cite{wang2024qwen2}    & 77.4    & 87.2 & 95.1     & 64.4     & 49.3   \\
                     & LLaVA-OV~\cite{li2024llava}    & 65.2    & 78.9 & 84.6     & 40.0     & 35.0   \\
                     & LLaVA-Video~\cite{lin2023video} & 70.4    & 81.2 & 90.7     & 53.3     & 39.3  \\
            \bottomrule
            \end{tabular}
        }
        &
        % Second Table
        \resizebox{0.47\textwidth}{!}{
        \begin{tabular}{cc|c|cccc}
            \toprule
Size                 & Model       & Overall & LLM  & Semantic & Temporal & Others \\
\midrule
\multirow{3}{*}{7B}  & Qwen2-VL~\cite{wang2024qwen2}    & 81.3    & 88.7 & 90.9     & 70.0     & 54.1   \\
                     & LLaVA-OV~\cite{li2024llava}    & 80.3    & 89.8 & 91.1     & 65.7    & 48.2   \\
                     & LLaVA-Video~\cite{lin2023video} & 84.4    & 92.8 & 92.3     & 73.7     & 56.7   \\
                     \midrule
\multirow{3}{*}{72B} & Qwen2-VL~\cite{wang2024qwen2}    & 84.0    & 91.1 & 92.6     & 70.9     & 60.0   \\
                     & LLaVA-OV~\cite{li2024llava}    & 83.2    & 93.4 & 93.9    & 66.6    & 50.6   \\
                     & LLaVA-Video~\cite{lin2023video} & 85.4    & 94.0 & 94.7     & 73.7    & 56.6  \\
            \bottomrule
            \end{tabular}
        } \\

           \small (c) VideoMME~\cite{fu2024video} & \small (d) MLVU~\citep{mlvu} \\
        \resizebox{0.47\textwidth}{!}{
        \begin{tabular}{cc|c|cccc}
            \toprule
Size                 & Model       & Overall & LLM  & Semantic & Temporal & Others \\
\midrule
\multirow{3}{*}{7B}  & Qwen2-VL~\cite{wang2024qwen2}    & 60.6    & 77.8 & 78.4     & 36.7     & 31.1   \\
                     & LLaVA-OV~\cite{li2024llava}    & 59.0    & 76.3 & 76.8    & 37.2     & 28.2   \\
                     & LLaVA-Video~\cite{lin2023video} & 63.9    & 79.3 & 82.0     & 42.6     & 34.7   \\
                     \midrule
\multirow{3}{*}{72B} & Qwen2-VL~\cite{wang2024qwen2}    & 68.2    & 86.8 & 86.3     & 49.6     & 33.8   \\
                     & LLaVA-OV~\cite{li2024llava}    & 68.7    & 87.2 & 86.3     & 52.6     & 33.6   \\
                     & LLaVA-Video~\cite{lin2023video} & 70.8    & 88.1 & 88.9     & 51.8     & 38.1  \\
            \bottomrule
            \end{tabular}
        }
        &
        % Second Table
        \resizebox{0.47\textwidth}{!}{
        \begin{tabular}{cc|c|cccc}
            \toprule
Size                 & Model       & Overall & LLM  & Semantic & Temporal & Others \\
\midrule
\multirow{3}{*}{7B}  & Qwen2-VL~\cite{wang2024qwen2}    & 62.5    & 77.8 & 79.5    & 43.6    & 37.4   \\
                     & LLaVA-OV~\cite{li2024llava}    & 65.2    & 77.1 & 88.0     & 47.5    & 36.1   \\
                     & LLaVA-Video~\cite{lin2023video} & 63.7    & 77.8 & 83.1     & 49.6     & 33.6   \\
                     \midrule
\multirow{3}{*}{72B} & Qwen2-VL~\cite{wang2024qwen2}    & 67.9    & 81.8 & 85.4     & 52.5     & 41.4   \\
                     & LLaVA-OV~\cite{li2024llava}    & 74.2    & 88.1 & 92.5    & 62.7    & 44.0  \\
                     & LLaVA-Video~\cite{lin2023video} & 74.2    & 87.4 & 92.5     & 64.0    & 43.8  \\
            \bottomrule
            \end{tabular}
        } \\

        \small (e) LongVideoBench~\cite{lvbench} & \small (f) PerceptionTest~\citep{perceptiontest} \\
        \resizebox{0.47\textwidth}{!}{
        \begin{tabular}{cc|c|cccc}
            \toprule
Size                 & Model       & Overall & LLM  & Semantic & Temporal & Others \\
\midrule
\multirow{3}{*}{7B}  & Qwen2-VL~\cite{wang2024qwen2}    & 52.8   & 74.4 & 70.5     & 42.6     & 27.6   \\
                     & LLaVA-OV~\cite{li2024llava}    & 58.9    & 79.1 & 82.1    & 49.8    & 30.2   \\
                     & LLaVA-Video~\cite{lin2023video} & 59.8    & 81.3 & 84.4     & 49.8     & 29.7   \\
                     \midrule
\multirow{3}{*}{72B} & Qwen2-VL~\cite{wang2024qwen2}    & 58.0    & 82.4 & 76.0     & 46.4     & 30.9   \\
                     & LLaVA-OV~\cite{li2024llava}    & 59.8    & 87.3 & 84.4     & 49.8     & 32.9   \\
                     & LLaVA-Video~\cite{lin2023video} & 62.8    & 87.3 & 85.7     & 52.8     & 31.3  \\
            \bottomrule
            \end{tabular}
        }
        &
        % Second Table
        \resizebox{0.47\textwidth}{!}{
        \begin{tabular}{cc|c|cccc}
            \toprule
Size                 & Model       & Overall & LLM  & Semantic & Temporal & Others \\
\midrule
\multirow{3}{*}{7B}  & Qwen2-VL~\cite{wang2024qwen2}    & 60.7    & 71.9 & 84.2    & 49.7    & 35.3   \\
                     & LLaVA-OV~\cite{li2024llava}    & 58.0    & 66.0 & 84.9     & 45.9    & 31.9   \\
                     & LLaVA-Video~\cite{lin2023video} & 68.3   & 75.4 & 87.9    & 60.8    & 47.8  \\
                     \midrule
\multirow{3}{*}{72B} & Qwen2-VL~\cite{wang2024qwen2}    & 68.1   & 77.7 & 92.1    & 62.7   & 40.5   \\
                     & LLaVA-OV~\cite{li2024llava}    & 62.5    & 75.6 & 89.8    & 50.6    & 32.8  \\
                     & LLaVA-Video~\cite{lin2023video} & 69.6    & 76.0 & 92.1     & 61.2    & 47.0  \\
            \bottomrule
            \end{tabular}
        } \\

    \end{tabular}
% \vspace{0.2cm}

    \label{tab:main}
\end{table}

\subsection{Benchmarking Public Models Under VBenchComp Categorization}

Table~\ref{tab:main} benchmarks recent public video-language models under our proposed VBenchComp framework, which categorizes questions into LLM-answerable, Semantic, and Temporal types. This fine-grained categorization provides a more diagnostic view of model capabilities compared to a single overall score. As shown in Table~\ref{tab:main}(a), Qwen2-VL-7B slightly outperforms LLaVA-Video-7B in terms of the traditional overall score on Egoschema. However, this superficial advantage is misleading. A breakdown of the scores shows that the performance gain is almost entirely due to LLM-answerable questions that do not require visual or temporal understanding. However, the two models perform similarly on Semantic questions, and Qwen2-VL-7B even lags behind on Temporal questions, which indicates a weaker grasp of fine-grained video temporal understanding. These findings suggest that Qwen2-VL-7B's advantage is largely attributable to its stronger language model backbone, rather than superior visual or temporal reasoning. In contrast, LLaVA-Video-7B, though slightly behind overall, demonstrates more balanced capabilities across semantic and temporal dimensions.

Interestingly, the comparison flips in VideoMME (Table~\ref{tab:main}(c)), where LLaVA-Video-7B outperforms Qwen2-VL-7B not just overall, but more meaningfully across both vision-dependent axes. While the two models perform similarly on LLM-answerable questions, LLaVA-Video-7B achieves notably higher scores on both Semantic (82.0 vs. 78.4) and Temporal (42.6 vs. 36.7) categories. This demonstrates that LLaVA-Video-7B possesses stronger visual and temporal understanding, reinforcing the claim that strong language knowledge alone are insufficient for robust video understanding. 

These results collectively demonstrate a core limitation of traditional evaluation: a single overall score fails to capture specific model strengths and weaknesses. Only through our VBenchComp categorization can we identify crucial gaps in  semantic or temporal understanding that would otherwise be masked. This insight is not only critical for fair benchmarking but also for guiding the development of next-generation video LLMs, where improvement must go beyond language modeling and target true temporal understanding.

\subsection{VBenchComp Score: Fewer Questions, Deeper Video Understanding}

\begin{figure}[htbp]
    \centering
    \begin{tabular}{ccc}
        \includegraphics[width=0.32\textwidth]{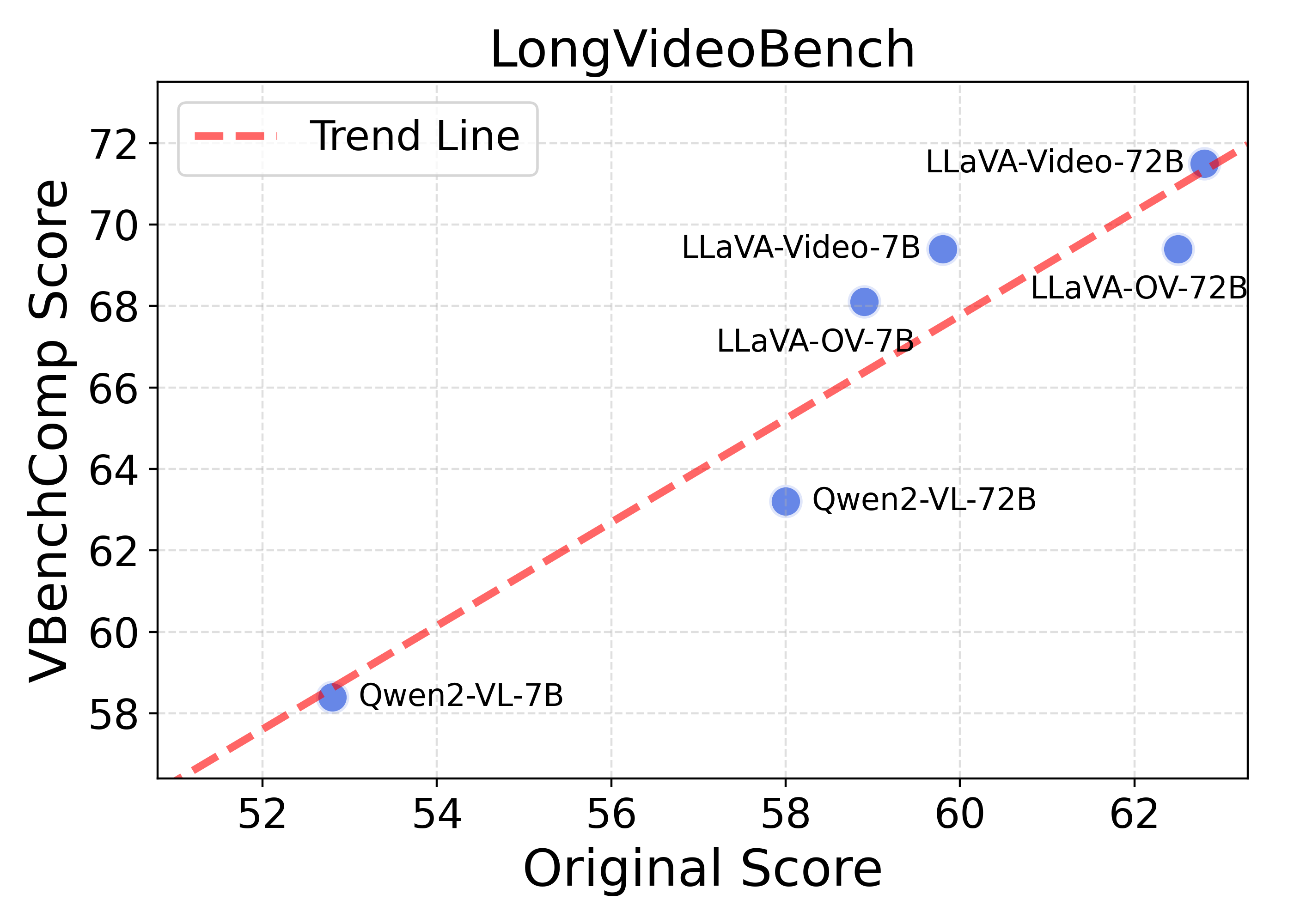} & \hspace{-0.5cm}
        \includegraphics[width=0.32\textwidth]{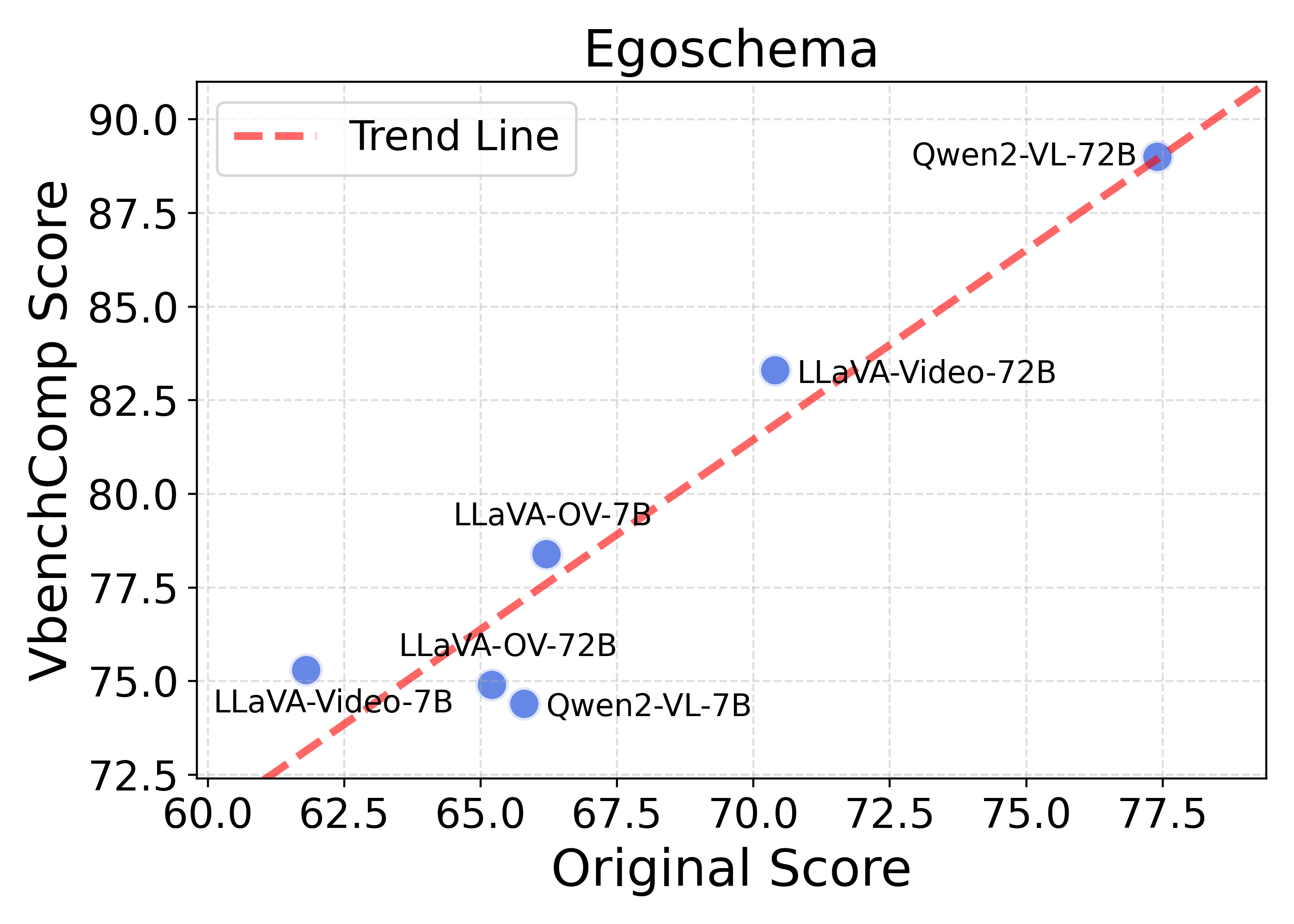} & \hspace{-0.5cm}
        \includegraphics[width=0.32\textwidth]{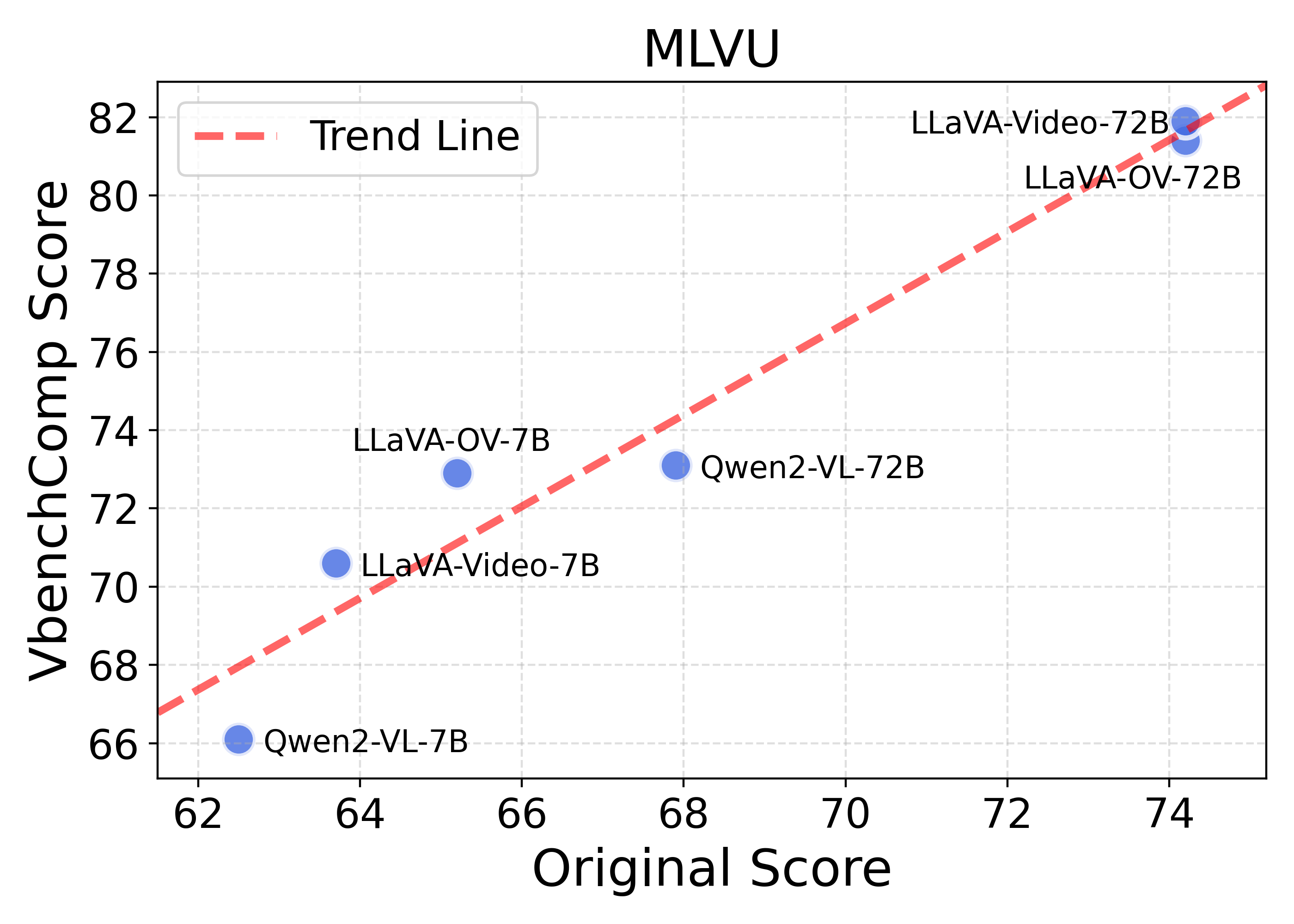} \\
        \includegraphics[width=0.32\textwidth]{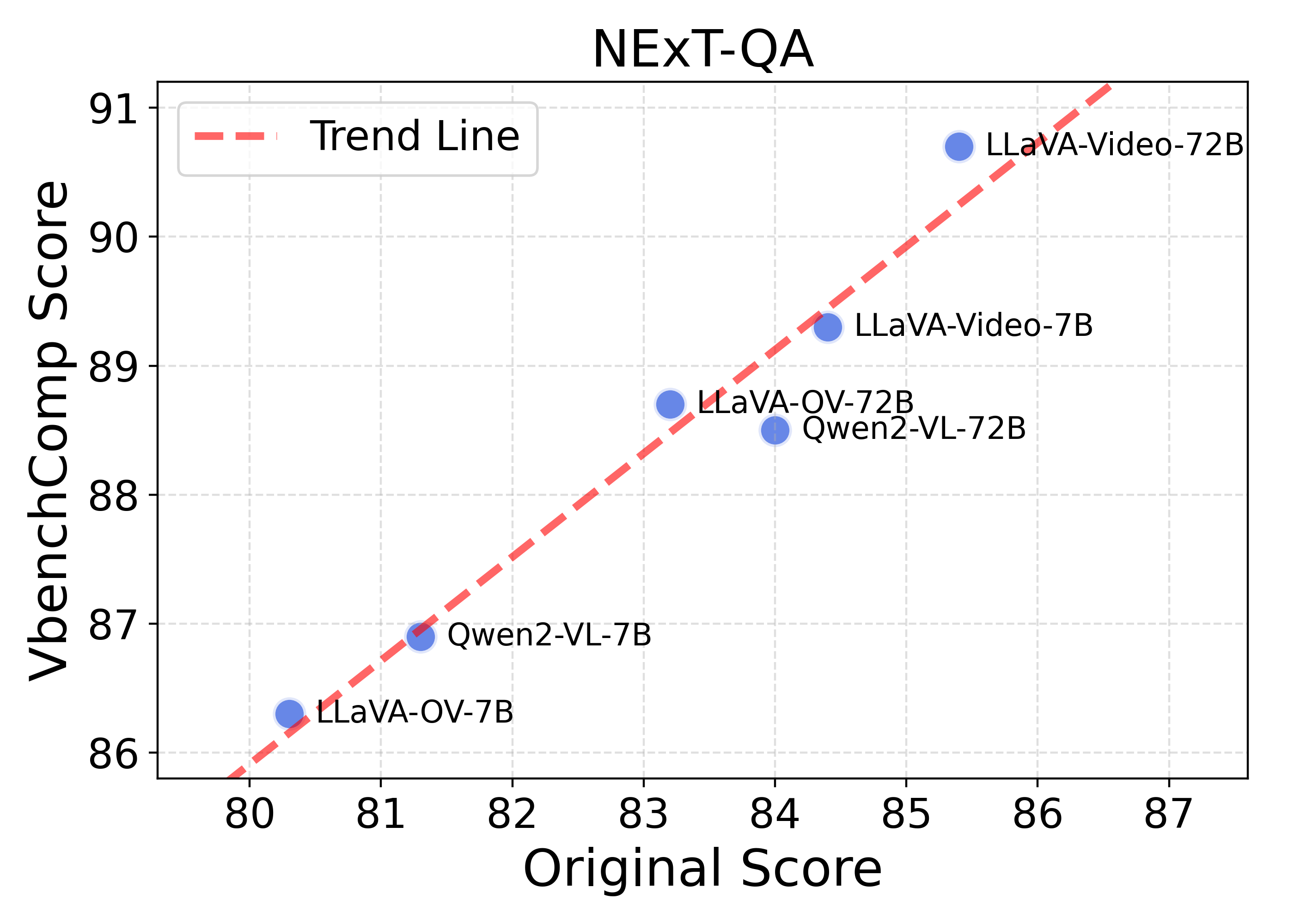} & \hspace{-0.5cm}
        \includegraphics[width=0.32\textwidth]{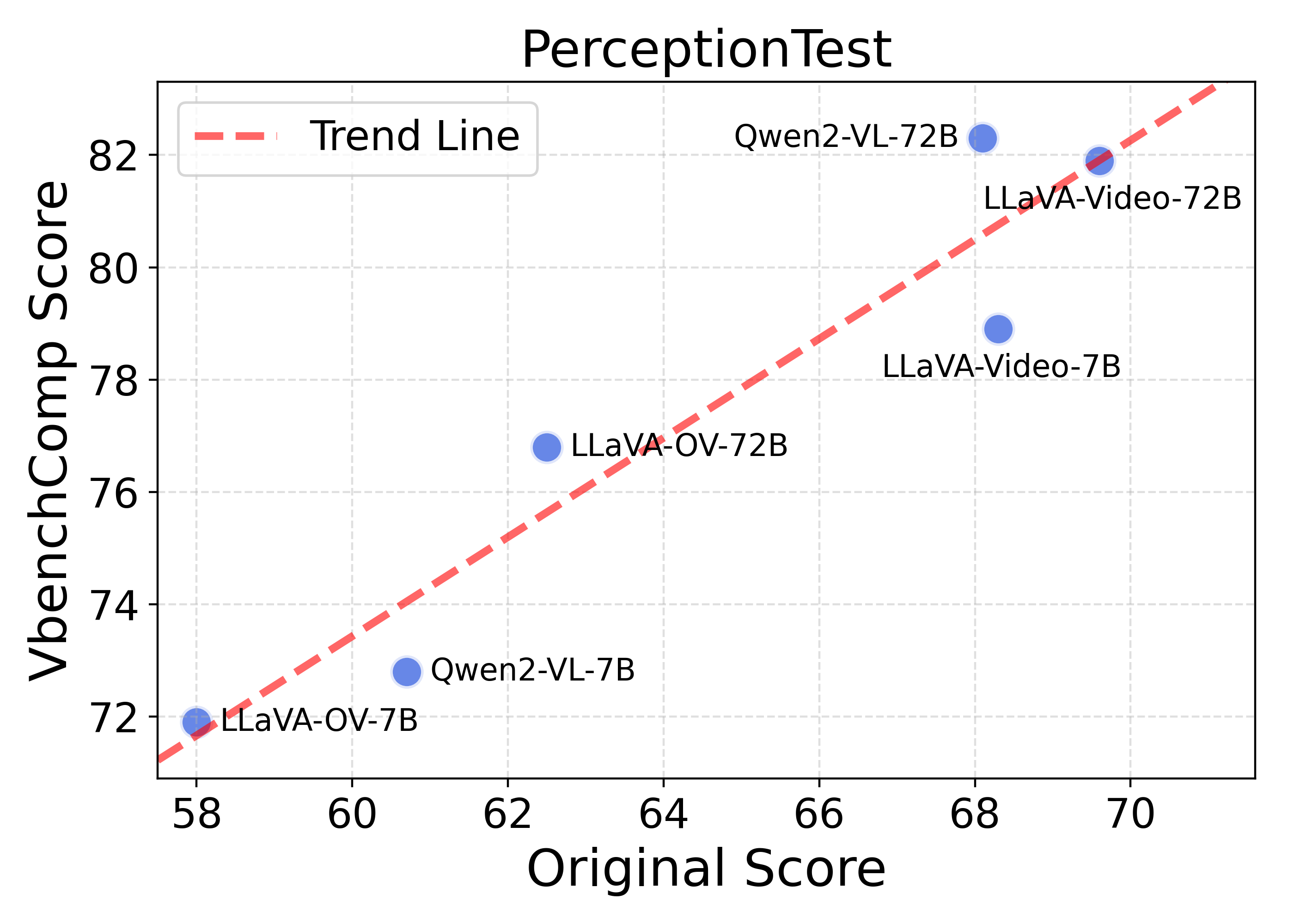} & \hspace{-0.5cm}
        \includegraphics[width=0.32\textwidth]{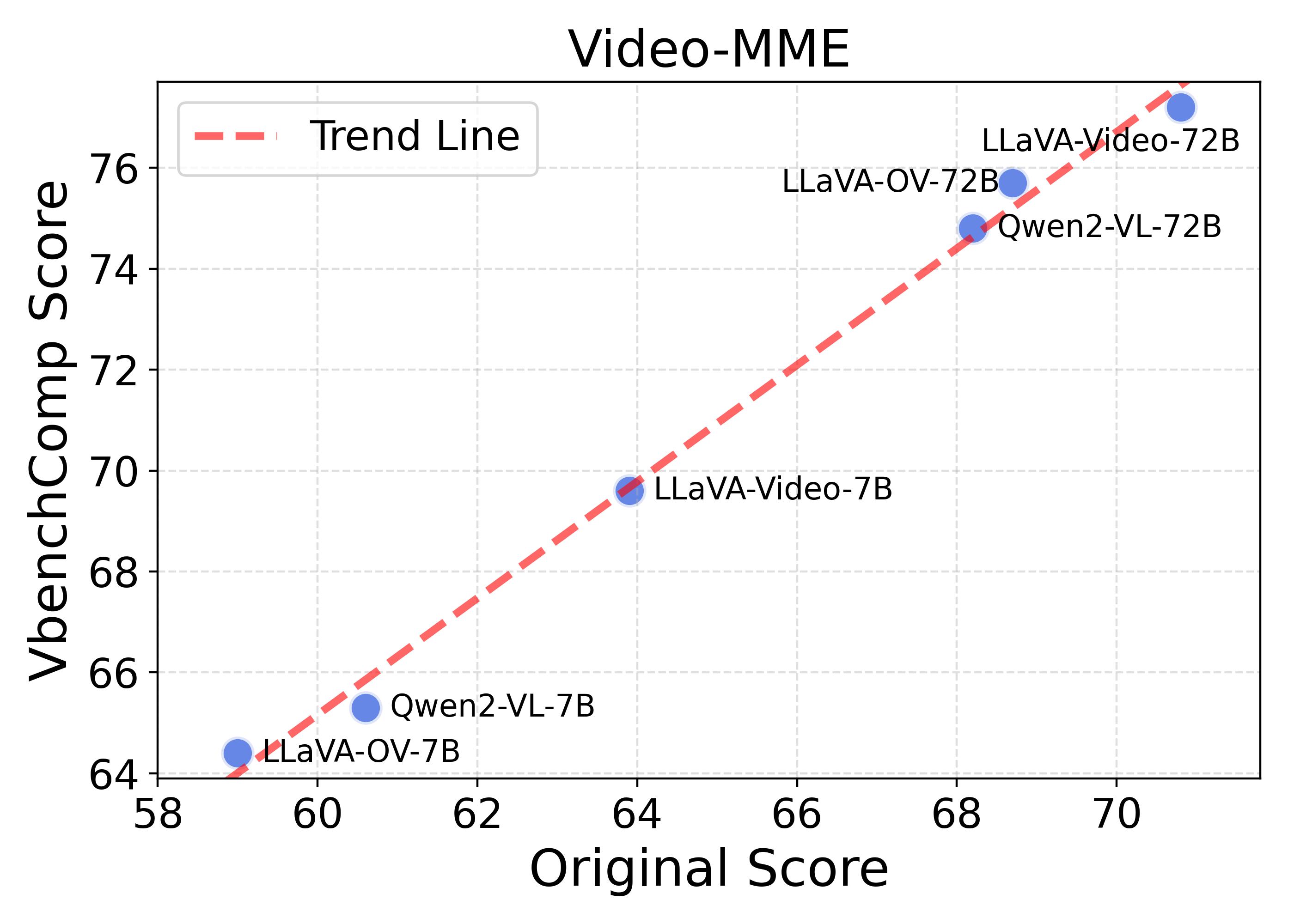} \\
    \end{tabular}
    \caption{VBenchComp scores are aligned with the original scores but they can better evaluate the overall video LLM performance with less questions. The temporal video understanding capability of models under the trend line can be potentially over-estimated in the original benchmarks.}
    \label{fig:cor}
    \vspace{-0.1cm}
\end{figure}

Based on the above analysis, we retain only the Semantic and Temporal questions from each benchmark to compute a focused evaluation score, denoted as the VBenchComp score. The results across models are shown in Figure~\ref{fig:cor}. Despite removing nearly 50\% of the original questions (as detailed in Table~\ref{tab: composition}), the model rankings remain highly consistent with those based on the original scores. This strong correlation indicates that Semantic and Temporal questions alone are sufficient to preserve the discriminative power of the benchmark. It further suggests that many of the remaining questions may be redundant or less critical for evaluating core model capabilities, and that VBenchComp can serve as a more focused yet reliable metric for model comparison.

\section{Discussion}

VBenchComp provides a structured and interpretable framework for dissecting the capabilities of video LLMs, highlighting whether models rely on language priors, static semantics, or genuine temporal reasoning. This diagnostic lens not only clarifies what current benchmarks actually measure, but also helps researchers identify blind spots in model behavior. However, our approach is not without limitations. First, while our categorization pipeline is automated and scalable, it heavily relies on GPT-4o and Gemini, which may introduce biases. Second, our core benchmark subset, while compute-efficient and representative in aggregate, may omit edge cases that appear in the full benchmark suite. Finally, VBenchComp focuses primarily on question-answering tasks; generalizing this framework to other video understanding tasks like captioning, retrieval, or grounding remains an important avenue for future work.

\section*{Acknowledgments}
We thank Mingze Xu, Junting Pan, Wentao Wu, Dongxu Li, Haotian Zhang, and Zhe Gan for their great help, feedback, and fruitful discussions. 

{
    \small
    \bibliographystyle{unsrt}
    \bibliography{neurips_2025}
}

\end{document}